# A Review of Research on Devnagari Character Recognition

Vikas J Dongre        Vijay H Mankar
Department of Electronics & Telecommunication,
Government Polytechnic, Nagpur, India

## ABSTRACT

English Character Recognition (CR) has been extensively studied in the last half century and progressed to a level, sufficient to produce technology driven applications. But same is not the case for Indian languages which are complicated in terms of structure and computations. Rapidly growing computational power may enable the implementation of Indic CR methodologies. Digital document processing is gaining popularity for application to office and library automation, bank and postal services, publishing houses and communication technology. Devnagari being the national language of India, spoken by more than 500 million people, should be given special attention so that document retrieval and analysis of rich ancient and modern Indian literature can be effectively done. This article is intended to serve as a guide and update for the readers, working in the Devnagari Optical Character Recognition (DOCR) area. An overview of DOCR systems is presented and the available DOCR techniques are reviewed. The current status of DOCR is discussed and directions for future research are suggested.

## 1. INTRODUCTION

Machine simulation of human functions has been a challenging research field since the advent of digital computers. In some areas, which require certain amount of intelligence, such as number crunching or chess playing, tremendous improvements are achieved. On the other hand, humans still outperform even the most powerful computers in the relatively routine functions such as vision. Machine simulation of human reading is one of these areas, which has been the subject of intensive research for the last three decades, yet it is still far from the final frontier.

The study investigates the direction of the Devnagari Optical Character Recognition research (DOCR), analyzing the limitations of methodologies for the systems which can be classified based upon two major criteria: the data acquisition process (on-line or off-line) and the text type (machine-printed or hand-written). No matter which class the problem belongs, in general there are five major stages in the DOCR problem:

1. Pre-processing, 2. Segmentation. 3. Feature Extraction, 4.Recognition, 5. Post processing.

The paper is arranged to review the DOCR methodologies with respect to the stages of the CR systems, rather than surveying the complete solutions. Although the off-line and on-line character recognition techniques have different approaches, they share a lot of common problems and solutions. Since it is relatively more complex and requires more research compared to on-line and machine-printed recognition, off-line handwritten character recognition is selected as a focus of attention in this article.

Handwriting Recognition Technology has been improving much under the purview of pattern recognition and image processing since a few decades. Hence various soft computing methods involved in other types of pattern and image recognition can as well be used for DOCR.

Seminal and comprehensive work in DOCR is carried out by R.M.K. Sinha and V. Bansal, [1-7]. A general Review of Statistical Pattern Recognition can also be found in [8-11]. These can be taken as good starting point to reach the recent studies in various types and applications of the DOCR problem. An excellent overview of document analysis can also be found in [12].

After describing the features of Devnagari language in Section 2, Image Pre-processing is discussed in Section 3, Segmentation is discussed in section 4, Feature Extraction types discussed in section 5, Character Classification is discussed in Section 6. Post processing is discussed in section 7. Finally, current research scenario and future research directions are discussed in Section 8.

## 2. FEATURES OF DEVNAGARI SCRIPT

India is a multi-lingual and multi-script country comprising of eighteen official languages. One of the defining aspects of Indian script is the repertoire of sounds it has to support. Because there is typically a letter for each of the phonemes in Indian languages, the alphabet set tends to be quite large. Most of the Indian languages originated from Bramhi script. These scripts are used for two distinct major linguistic groups, Indo-European languages in the north, and Dravidian languages in the south [16].

Devnagari is the most popular script in India. It has 11 vowels and 33 consonants. They are called basic characters. Vowels can be written as independent letters, or by using a variety of diacritical marks which are written above, below, before or after the consonant they belong to. When vowels are written in this way they are known as *modifiers* and the characters so formed are called *conjuncts*. Sometimes two or more consonants can combine and take new shapes. These new shape clusters are known as *compound characters*.

These types of basic characters, compound characters and modifiers are present not only in Devnagari but also in other scripts. Hindi, the national language of India, is written in the Devnagari script. Devnagari is also used for writing Marathi, Sanskrit and Nepali. Moreover, Hindi is the third most popular language in the world [11]. A sample of Devnagari character set is provided in table 1 to 6.

**Table 1: Vowels and Corresponding Modifiers**.

| Vowels: | अ | आ | इ | ई | उ | ऊ | ऋ | ए | ऐ | ओ | औ |
|---|---|---|---|---|---|---|---|---|---|---|---|
| Modifiers: |  | ा | ि | ी | ु | ू | ृ | े | ै | ो | ौ |

**Table 2: Consonants**

| क | ख | ग | घ | ङ | च | छ | ज | झ | ञ | ट |
|---|---|---|---|---|---|---|---|---|---|---|
| ठ | ड | ढ | ण | त | थ | द | ध | न | प | फ |
| ब | भ | म | य | र | ल | व | श | ष | स | ह |

**Table 3: Half Form of Consonants with Vertical Bar.**

| क् |रव् | ग् | घ् |  | च् |  | ज् | झ् | ञ् |
|---|---|---|---|---|---|---|---|---|---|
|  |  |  | ण् | त् | थ् |  | ध् | न् | प् | फ् |
| ब् | भ् | म् | य् |  | ल् | व् | श् | ष् | स् |

**Table 4: Examples of Combination of Half-Consonant and Consonant.**

| क क्क | क ल्क | घ न्घ | च ञ्च | ज च्ज | त न्त | प त्प | प ल्प |
|---|---|---|---|---|---|---|---|
| व व्व | भ न्भ | म ल्म | ल ल्ल | श न्श | श व्श | श ल्श | स न्स |

**Table 5: Examples of Special Combination of Half-Consonant and Consonant.**

| क्ष | ज्ञ | ट्ट | ट्ठ | त्र | द्द |
|---|---|---|---|---|---|
| द्ध | द्व | द्र | श्र | द्भ | द्य |

**Table 6: Special Symbols**

| क़ | ख़ | ग़ | ज़ | फ़ | ड़ | ढ़ | . | ० | : | । | ऽ | ं |
|---|---|---|---|---|---|---|---|---|---|---|---|---|

All the characters have a horizontal line at the upper part, known as *Shirorekha* or headline. No English character has such characteristic and so it can be taken as a distinguishable feature to extract English from these scripts. In continuous handwriting, from left to right direction, the shirorekha of one character joins with the shirorekha of the previous or next character of the same word. In this fashion, multiple characters and modified shapes in a word appear as a single connected component joined through the common shirorekha. All the characters and modified shapes in a word appear to hang from the hypothetical shirorekha of the word. Also in Devnagari there are vowels, consonants, vowel modifiers and compound characters, numerals. Moreover, there are many similar shaped characters. All these variations make DOCR, a challenging problem [13].

## 3. IMAGE PREPROCESSING

Data in a paper document are usually captured by optical scanning and stored in a file of picture elements, called pixels. These pixels may have values: OFF (0) or ON (1) for binary images, 0– 255 for gray-scale images, and 3 channels of 0–255 colour values for colour images. This collected raw data must be further analyzed to get useful information. Such processing includes the following:

### 3.1 Thresholding:
A grayscale or colour image is reduced to a binary image.

### 3.2 Noise reduction:
The noise, introduced by the optical scanning device or the writing instrument, causes disconnected line segments, bumps and gaps in lines, filled loops etc. The distortion including local variations, rounding of corners, dilation and erosion, is also a problem. Prior to the character recognition, it is necessary to eliminate these imperfections [23-24].

### 3.3 Skew Detection and Correction:
Handwritten document may originally be skewed or skewness may introduce in document scanning process. This effect is unintentional in many real cases, and it should be eliminated because it dramatically reduces the accuracy of the subsequent processes, such as segmentation and classification. Skewed lines are made horizontal by calculating skew angle and making proper correction in the raw image [14], [21-22].

### 3.4 Size Normalization:
Each segmented character is normalized to fit within suitable matrix like 32x32 or 64x64 so that all characters have same data size [14].

### 3.5 Thinning:
The boundary detection of image is done to enable easier subsequent detection of pertinent features and objects of interest (see fig.1 (d)). Various standard functions are now available in MATLAB for above operations [67].

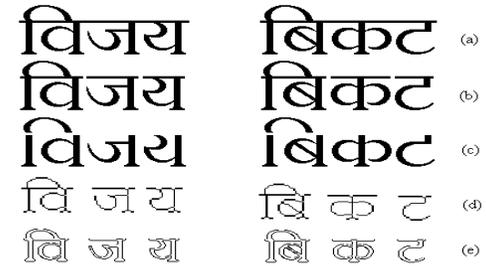

**Figure 1: Preprocessed Images (a) Original, (b) segmented (c) Shirorekha removed (d) Thinned (e) image edging**

## 4. SEGMENTATION
It is one the most important process that decides the success of character recognition technique. It is used to decompose an image of a sequence of characters into sub images of individual symbols by segmenting lines and words [17], [43]. Devnagari words can further be splitted to individual character for classification and recognition by removing Shirorekha (see fig.1 (c)). Various vowel modifiers can be separated for structural feature extractions [18-20].

## 5. FEATURE EXTRACTION
Feature extraction and selection can be defined as extracting the most representative information from the raw data, which minimizes the within class pattern variability while enhancing the between class pattern variability. For this purpose, a set of features are extracted for each class that helps distinguish it from other classes, while remaining invariant to characteristic differences within the class [25]. A good survey on feature extraction methods for character recognition can be found in [26]. Various feature extraction methods are classified in three major groups:
1. Global Transformation and Series Expansion
2. Statistical Features
3. Geometrical and Topological Features

### 5.1 Global Transformation and Series Expansion
A continuous signal generally contains more information than needs to be represented for the purpose of classification. One way to represent a signal is by a linear combination of a series of simple well-defined functions. The coefficients of the linear combination provide a compact encoding known as transformation or/and series expansion. Deformations like translation and rotation are invariant under global transformation and series expansion. Common transform and series expansion methods used in the CR field are:

*5.1.1 Fourier Transforms*: The general procedure is to choose magnitude spectrum of the measurement vector as the features in an n-dimensional Euclidean space. One of the most attractive properties of the Fourier Transform is the ability to recognize the position-shifted characters, when it observes the magnitude spectrum and ignores the phase [24]. Fourier Transforms has been applied to Devnagari OCR in many ways [27].

*5.1.2 Gabor Transform:* It is a variation of the windowed Fourier Transform. In this case, the window used is not a discrete size, but is defined by a Gaussian function

*5.1.3 Wavelets*: Wavelet transformation is a series expansion technique that allows us to represent the signal at different levels of resolution. The segments of document image, which may correspond to letters or words, are represented by wavelet coefficients, corresponding to various levels of resolution. These coefficients are then fed to a classifier for recognition [28-29].

*5.1.4 Moments:* Moments, such as central moments, Legendre moments, Zernike moments, form a compact representation of the original document image that make the process of recognizing an object scale, translation, and rotation invariant [30-32], [45]. Moments are considered as series expansion representation, since the original image can be completely reconstructed from the moment coefficients.

*5.1.5 Karhunen-Loeve Expansion:* It is an eigen-vector analysis, which attempts to reduce the dimension of the feature set by creating new features that are linear combinations of the original ones. It is the only optimal transform in terms of information compression. Karhunen-Loeve expansion is used in several pattern recognition problems such as face recognition. Since it requires computationally complex algorithms, the use of Karhunen-Loeve features in CR problems is not widespread. However, by the increase of the computational power, it is gaining importance.

### 5.2 Statistical Features
Representation of a document image by statistical distribution of points takes care of style variations to some extent. Although this type of representation does not allow the reconstruction of the original image, it is used for reducing the dimension of the feature set providing high speed and low complexity. The major statistical features mentioned below are used for character representation

*5.2.1 Zoning:* The frame containing the character is divided into several overlapping or non-overlapping zones. The densities of the points or some features in different regions are analyzed [33].

*5.2.2 Crossings and Distances:* A popular statistical feature is the number of crossing of a contour by a line segment in a specified direction. The character frame is partitioned into a set of regions in various directions and then features of each region are extracted.

*5.2.3 Projections*: Characters can be represented by projecting the pixel gray values onto lines in various directions. This representation creates one-dimensional signal from a two dimensional image, which can be used to represent the character image [34].

### 5.3 Geometrical and Topological Features
Various global and local properties of characters can be represented by geometrical and topological features with high tolerance to distortions and style variations. This type of representation may also, encode some knowledge about the structure of the object or may provide some knowledge as to what sort of components make up that object. Various topological and geometrical representations can be grouped in four categories:

*5.3.1 Extracting and Counting Topological Structures:* In this category, lines, curves, splines, extreme points, maxima and minima, cups above and below a threshold, openings, to the right, left, up and down, cross (X) points, branch (T) points, line ends (J), loops (O), direction of a stroke from a special point, inflection between two points, isolated dots, a bend between two points, horizontal curves at top or bottom, straight strokes between two points, ascending, descending and middle strokes and relations among the stroke that make up a character are considered as features [26], [35].

*5.3.2 Measuring and Approximating the Geometrical Properties:* In this category, the characters are represented by the measurement of the geometrical quantities such as, the ratio between width and height of the bounding box of a character, the relative distance between the last point and the last y-min, the relative horizontal and vertical distances between first and last points, distance between two points, comparative lengths between two strokes, width of a stroke, upper and lower masses of words, word length curvature or change in the curvature[15], [35-36], [42].

*5.3.2 Coding:* One of the most popular coding schemes is Freeman's chain code. This coding is essentially obtained by mapping the strokes of a character into a 2-dimensional parameter space, which is made up of codes. There are many versions of chain coding. The character frame is divided to left-right sliding window and each region is coded by the chain code [36-39].

*5.3.4 Graphs and Trees:* Words or characters are first partitioned into a set of topological primitives, such as strokes, holes, cross points etc. Then, these primitives are represented using attributed or relational graphs. Image is represented either by graphs coordinates of the character shape or by an abstract representation with nodes corresponding to the strokes and edges corresponding to the relationships between the strokes. Trees can also be used to represent the words or characters with a set of features, which has a hierarchical relation [40-41], [65].

## 6. CHARACTER CLASSIFICATION
OCR systems extensively use the methodologies of pattern recognition, which assigns an unknown sample to a predefined class. Numerous techniques for OCR are investigated by the researchers. A good survey on feature extraction and classification methods for Devnagari character recognition can be found in [15], [24]. OCR classification techniques can be classified as follows.
1. Template Matching.
2. Statistical Techniques.
3. Neural Networks.
4. Support Vector Machine (SVM) algorithms.
5. Combination classifier.

The above approaches are neither necessarily independent nor disjoint from each other. Occasionally, a CR technique in one approach can also be considered to be a member of other approaches.

### 6.1 Template Matching
This is the simplest way of character recognition, based on matching the stored prototypes against the character or word to be recognized. The matching operation determines the degree of similarity between two vectors (group of pixels, shapes, curvature etc.) A gray-level or binary input character is compared to a standard set of stored prototypes. According to a similarity measure (e.g.: Euclidean, Mahalanobis, Jaccard or Yule similarity measures etc). A template matcher can combine multiple information sources, including match strength and k-nearest neighbor measurements from different metrics. The recognition rate of this method is very sensitive to noise and image deformation. For improved classification Deformable Templates and Elastic Matching are used [44-45].

### 6.2 Statistical Techniques
Statistical decision theory is concerned with statistical decision functions and a set of optimality criteria, which maximizes the probability of the observed pattern given the model of a certain class. [41]. Statistical techniques are based on following assumptions:

a. Distribution of the feature set is Gaussian or in the worst case uniform,
b. There are sufficient statistics available for each class,
c. Given collection of images is able to extract a set of features which represents each distinct class of patterns.

The measurements taken from n-features of each word unit can be thought to represent an n-dimensional vector space and the vector, whose coordinates correspond to the measurements taken, represents the original word unit. The major statistical methods, applied in the OCR field are Nearest Neighbor (NN) [46-47], Likelihood or Bayes classifier [49], Clustering Analysis [52], Hidden Markov Modeling (HMM) [36], Fuzzy Set Reasoning [50-51], [65], Quadratic classifier [53].

### 6.3 Neural Networks
Character classification problem is related to heuristic logic as human beings can recognize characters and documents by their learning and experience. Hence neural networks which are more or less heuristic in nature are extremely suitable for this kind of problem. Various types of neural networks are used for OCR classification.

A neural network is a computing architecture that consists of massively parallel interconnection of adaptive 'neural' processors. Because of its parallel nature, it can perform computations at a higher rate compared to the classical techniques. Because of its adaptive nature, it can adapt to changes in the data and learn the characteristics of input signal [12]. Output from one node is fed to another one in the network and the final decision depends on the complex interaction of all nodes.

Several approaches exist for training of neural networks viz. error correction, Boltzman, Hebbian and competitive learning. They cover binary and continuous valued input, as well as supervised and unsupervised learning.

Neural network architectures can be classified as, feed-forward and feedback (recurrent) networks. The most common neural networks used in the OCR systems are the multilayer perceptron (MLP) of the feed forward networks and the Kohonen's Self Organizing Map (SOM) of the feedback networks. One of the interesting characteristics of MLP is that in addition to classifying an input pattern, they also provide a confidence in the classification [9]. These confidence values may be used for rejecting a test pattern in case of doubt. MLP is proposed by U. Bhattacharya *et al.* [46-47]. A detailed comparison of various NN classifiers is made by M. Egmont-Petersen [54]. He has shown that Feed-forward, perceptron higher order network, Neuro-fuzzy system are better suited for character recognition [51]. K. Y. Rajput *et al.* [57] used back propagation type NN classifier. Genetic algorithm based feature selection and classification along with fusion of NN and Fuzzy logic is reported in English [36], [64] but no any work is reported for Indian languages.

### 6.4 Support Vector Machine Classifier
It is primarily a two-class classifier. Width of the margin between the classes is the optimization criterion, i.e., the empty area around the decision boundary defined by the distance to the nearest training patterns [9]. These patterns, called support vectors, finally define the classification function. Their number is minimized by maximizing the margin. The support vectors replace the prototypes with the main difference between SVM and traditional template matching techniques is that they characterize the classes by a decision boundary. Moreover, this decision boundary is not just defined by the minimum distance function, but by a more general possibly nonlinear, combination of these distances. Many researchers used SVM successfully viz. Sandhya Arora *et al*. [46], C. V. Jawahar *et al*. [55], Umapada Pal *et al.* [56].

## 6.5 Combination Classifier

Various classification methods have their own superiorities and weaknesses. Hence many times multiple classifiers are combined together to solve a given classification problem. Different classifiers trained on the same data may not only differ in their global performances, but they also may show strong local differences. Each classifier may have its own region in the feature space where it performs the best. Some classifiers such as neural networks show different results with different initializations due to the randomness inherent in the training procedure. Instead of selecting the best network and discarding the others, one can combine various networks, thereby taking advantage of all the attempts to learn from the data [9].

In summary, we may have different feature sets, different training sets, different classification methods or different training sessions, all resulting in a set of classifiers, whose outputs may be combined, with the hope of improving the overall classification accuracy [9]. If this set of classifiers is fixed, the problem focuses on the combination function. It is also possible to use a fixed combiner and optimize the set of input classifiers. A typical combination scheme consists of a set of individual classifiers and a combiner which combines the results of the individual classifiers to make the final decision.

Various schemes for combining multiple classifiers can be grouped into three main categories according to their architecture: 1) parallel, 2) cascading (or serial combination) and 3) hierarchical (tree-like)

### *6.5.1 Selection and Training of Individual Classifiers*

A classifier combination is especially useful if the individual classifiers are largely independent. If this is not already guaranteed by the use of different training sets, various resampling techniques like rotation and bootstrapping may be used to artificially create such differences for improving the classification rate [9].

### *6.5.2 Combiner*

After individual classifiers have been selected, they need to be combined together by a module, called the combiner. Various combiners can be distinguished from each other in their trainability, adaptivity, and requirement on the output of individual classifiers. Combiners, such as voting, averaging (or sum), and Borda count are static, with no training required, while others are trainable. The trainable combiners may lead to a better improvement than static combiners at the cost of additional training as well as the requirement of additional training data. Some combination schemes are adaptive in the sense that the combiner evaluates (or weighs) the decisions of individual classifiers depending on the input pattern. In contrast, nonadaptive combiners treat all the input patterns the same.

Some combination classifiers used in Indian scripts are ANN and HMM [59], K-Means and SVM [61], MLP and SVM [62], MLP and minimum edit [63], SVM and ANN [46], fuzzy neural network [51], NN, fuzzy logic and genetic algorithm [64]. Pavan Kumar [60] used five different classifiers (two HMM and three NN based) to obtain better accuracy.

**Table 7: Comparison of Numeral Results by Researchers.**

| S.N | Method proposed by | Data size | Accuracy obtained |
|---|---|---|---|
| 1 | R. Bajaj *et al*. [39] | 400 | 89.6% |
| 2 | R. J. Ramteke *et al*. [45] | 169 | 92.28% |
| 3 | U. Bhattacharya *et al*. [59] | 16273 | 95.64% |
| 4 | N. Sharma *et al*. [53] | 22,556 | 98.86% |

**Table 8: Comparison of Character Results by Researchers.**

| S.N | Method proposed by | Data size | Accuracy obtained |
|---|---|---|---|
| 1 | Kumar and Singh [66] | 200 | 80% |
| 2 | N. Sharma *et al*.[53] | 11270 | 80.36% |
| 4 | Sandhya Arora *et al*.[63] | 4900 | 92.80% |
| 5 | U. Pal *et al.* [15] | 36172 | 95.19% |

## 7. POSTPROCESSING

It is well known that humans read by context up to 60% for careless handwriting. While preprocessing tries to clean the document in a certain sense, it may remove important information, since the context information is not available at this stage. If the semantic information were available to a certain extent, it would contribute a lot to the accuracy of the CR stages. On the other hand, the entire OCR problem is for determining the context of the document image. Therefore the incorporation of context and shape information in all the stages of OCR systems is necessary for meaningful improvements in recognition rates. This is done in the postprocessing stage with a feedback to the early stages of OCR. The simplest way of incorporating the context information is the utilization of a dictionary for correcting the minor mistakes of the OCR systems. The basic idea is to spell check the OCR output and provide some alternatives for the outputs of the recognizer that do not take place in the dictionary. Research is underway for spelling checkers for Devnagari language. After the unknown character is recognized, it can be saved in text file for further editing or/and other applications using standard word processors.

## 8. FUTURE RESEARCH

Research and development in Indic language processing is a necessity for a highly multilingual, multiple-script country like India. Ministry of Information Technology of Government of India started a program on Technology Development for Indian Languages (TDIL: http://www.tdil.mit.gov.in) where language aspects are studied and developed. Another Government undertaking CDAC (Centre for Development of Advance Computing) is actively involved in development of Indian languages fonts, translators). Various hardware and software based language processors and language translators are developed by CDAC in collaboration with IIT Kanpur and indigenously (GIST, LIPI, ISM for word processing and Chitrankan software for offline character recognition). R M K Sinha of IIT Kanpur has been instrumental in the development of Indian language recognition and processing since the beginning. ISCII (Indian Scripts Standard Code for Information Interchange), the Indian standards for various languages was developed in 1988 by Indian Government. Also various Indic script symbols are incorporated in Unicode consortium (http://www.unicode.org/ charts/PDF/U0980.pdf). Research in Devnagari character is currently undergoing in leading institutes in IIT Kanpur, IIIT Hydrabad, ISI Kolkata and many others [8].

Researchers have investigated OCR for a number of Indian scripts: Devnagari, Tamil, Telugu, Bengali, and Kannada, Gurumukhi. However, most of this research has been confined to the identification of isolated characters rather than the script. Some systems used a statistical method; others were syntactic and/or heuristic-based. Unlike roman script, the Indic scripts are a composition of the constituent symbols in two dimensions. In conventional Research, first a word is segmented into its composite characters. Each composite character is then decomposed into the constituent symbols or the strokes (diacritic marks like *matra*) that are finally recognized. Holistic approaches circumvent the issues of segmentation ambiguity and character shape variability that are primary concerns for analytical approaches, and they may succeed on poorly written words where analytical methods fail to identify character content[14], [48]. A lot of research is still needed for word, sentence and document recognition, its semantics and lexicon. There is still a dearth of need to do the research in the area Devnagari character recognition.

## 9. CONCLUSION

Methods for treating the problem of Devnagari character recognition have developed remarkably in the last two decades. Still a lot of research is needed to tackle the challenges in DOCR so that commercially viable software solutions can be made available. It is hoped that this comprehensive discussion will provide insight into various concepts involved, and boost further advances in the area.

The difficulty of performing accurate recognition is determined by the nature of the material to be read and by its quality. Generally, misrecognition rates for unconstrained material increase progressively from machine print to handwritten writing. Methods of increasing sophistication are being pursued. Current research employs models not only of characters, but also words and phrases, and even entire documents. The powerful tools such as HMM, neural networks and their combinations are used. In order to have high reliability in character recognition, segmentation and classification have to be treated in an integrated manner to obtain more accuracy in complex cases. This paper has concentrated on an appreciation of principles and methods. Present work has not attempted to compare the effectiveness of various algorithms. It would be difficult to assess techniques separate from the systems for which they were developed. Unfortunately there is little experimental as well as standard handwritten character database available publicly for benchmarking the accuracy of various advanced techniques proposed in Devnagari character recognition [68].

The list of references to provide more detailed understanding of the approaches described is enlisted. We apologize to researchers whose important contributions may have been overlooked.

**Vikas. J Dongre** received B.E and M.E. in Electronics in

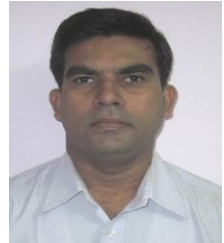

19991 and 1994 respectively. He served as lecturer in SSVPS engineering college Dhule, (M.S.) India from 1992 to 1994. He Joined Government Polytechnic Nagpur as Lecturer in 1994 where he is presently working as lecturer (selection grade). His areas of interests include Microcontrollers, embedded systems, image recognition, and innovative Laboratory practices.

**Vijay H. Mankar** received M. Tech. degree in Electronics

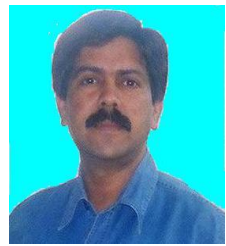

Engineering from VNIT, Nagpur University, India in 1995 and Ph.D. (Engg) from Jadavpur University, Kolkata, India in 2009 respectively. He has more than 16 years of teaching experience and presently working as a Lecturer (Selection Grade) in Government Polytechnic, Nagpur (MS), India. He has published more than 30 research papers in international conference and journals. His field of interest includes digital image processing, data hiding and watermarking.